\documentclass{article}
\usepackage{ijcai26}

\usepackage{times}
\usepackage{soul}
\usepackage{url}
\usepackage[hidelinks]{hyperref}
\usepackage[utf8]{inputenc}
\usepackage[small]{caption}
\usepackage{graphicx}
\usepackage{amsmath}
\usepackage{amssymb}
\usepackage{amsthm}
\usepackage{booktabs}
\usepackage{algorithm}
\usepackage{algorithmic}
\usepackage[switch]{lineno}
\usepackage{xspace}
\usepackage{placeins}

\newcommand{\evom}{\textsc{EVOM}\xspace}
\newcommand{\mles}{\textsc{MLES}\xspace}

\title{\evom: Agentic Meta-Evolution of Actor-Critic Architectures for Reinforcement Learning}

\author{
Boyun Zhang, Chao Wang\thanks{Corresponding author.}, Kai Wu
\affiliations
Xidian University
\emails
23009300248@stu.xidian.edu.cn, xiaofengxd@126.com, kwu@xidian.edu.cn
}

\begin{document}

\maketitle

\begin{abstract}
In actor-critic reinforcement learning, network architectures are typically manually designed. Automating this design is challenging because each candidate must be trained before evaluation, and the design space is open-ended. To address these challenges, we introduce \evom, an agentic meta-evolution framework for discovering high-performance actor-critic architectures. We frame architecture search as a bi-level optimization: an inner loop trains weights via the low-fidelity proximal policy optimization (PPO), while an outer loop drives meta-evolution by iteratively refining architecture programs. Crucially, this outer loop is powered by an LLM-based design agent that operates purely as an architecture designer, completely decoupled from policy execution and environment control. Experiments reveal that \evom outperforms the manually designed baseline, an LLM-guided random search, and the state-of-the-art LLM-guided programmatic policy search method \mles, delivering superior performance on Ant-v4 and HalfCheetah-v4. Ablation studies validate that both the meta-evolution loop and the LLM Design Agent are indispensable for final performance.
\end{abstract}

\section{Introduction}

The success of deep learning \cite{lecun2015deep} has been driven not only by algorithmic advances but also by innovations in network architecture. In supervised learning, this observation has motivated extensive research on neural architecture search (NAS) \cite{elsken2019neural,liu2021survey}, producing architectures that are competitive with or better than hand-designed alternatives across vision and language benchmarks. In reinforcement learning (RL) \cite{sutton1998reinforcement,li2024bridging}, however, automated architecture design has received less attention. Within actor-critic methods, such as proximal policy optimization (PPO) \cite{schulman2017proximal}, the networks used to instantiate the policy and value functions are typically treated as fixed implementation choices inherited from prior work. This convention may overlook an important design factor: actor-critic architectures can affect learning stability and final performance, motivating automated approaches to architecture discovery for actor-critic systems.

\begin{figure*}[t]
\centering
\includegraphics[width=\textwidth]{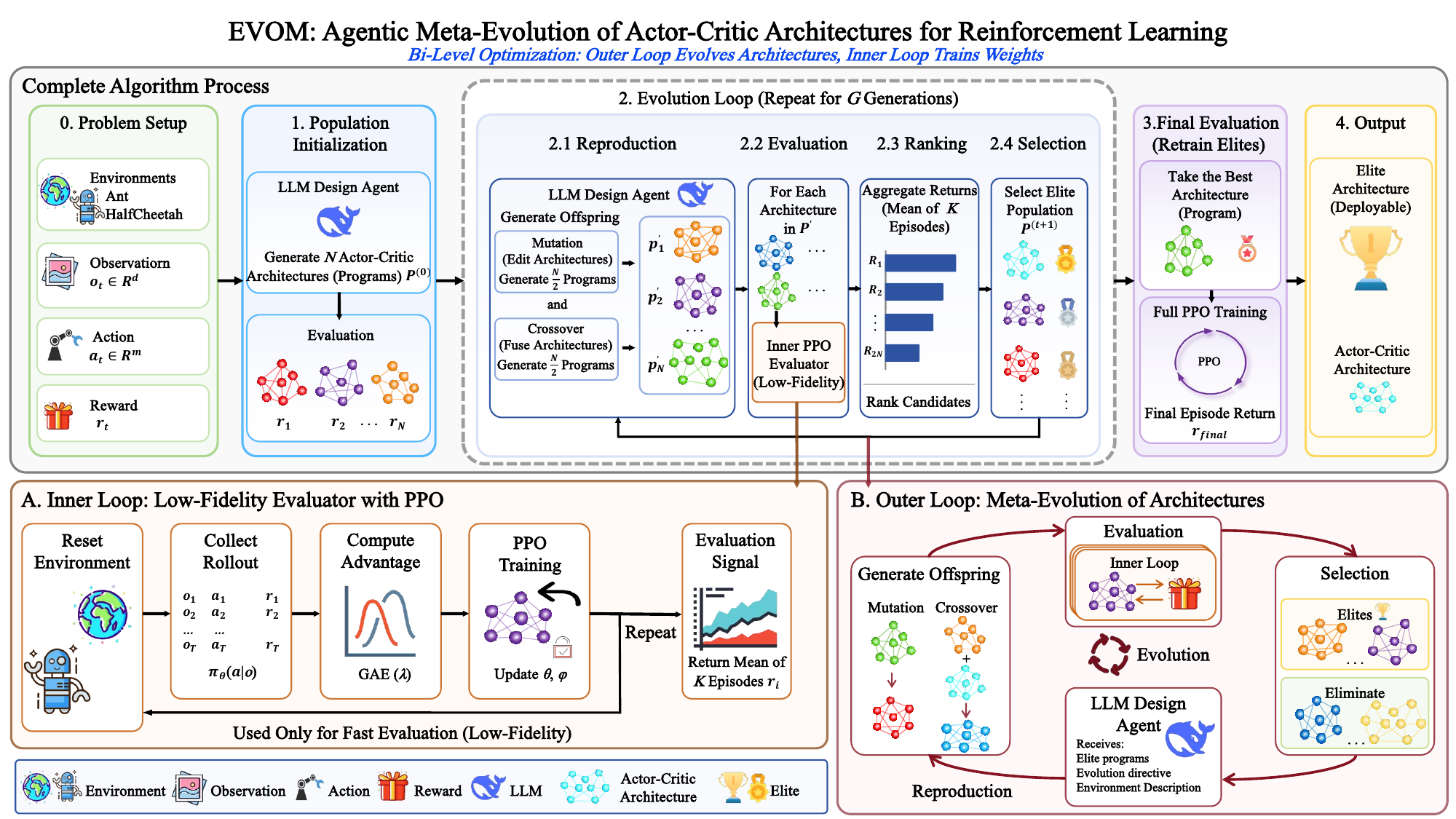}
\caption{Overview of \evom as bi-level optimization: Outer loop evolves architectures, and inner loop trains weights.}
\label{fig:overview}
\end{figure*}

Automated architecture discovery for actor-critic methods faces two main challenges. First, each candidate architecture must be trained before it can be evaluated, making full-budget comparison across many candidates computationally expensive. Second, actor-critic design is open-ended: policy heads, value heads, shared trunks, normalization layers, activation functions, and other modules can be composed in many ways, without a canonical fixed-topology search space. We propose \evom, an agentic meta-evolution framework that jointly addresses these challenges. A low-fidelity inner evaluation loop provides computationally tractable fitness estimates. At the same time, an LLM-based design agent generates and refines architectures as executable programs, reducing reliance on a manually predefined search space. As illustrated in Fig.~\ref{fig:overview}, \evom operates as a bi-level optimization: the outer loop evolves a population of architecture programs, and the inner loop evaluates each candidate by training its weights with PPO under a reduced budget.

\evom also differs from recent LLM-guided policy search methods such as \mles \cite{hu2025mles}, which synthesize executable controller programs that map observations directly to actions. In contrast, \evom generates trainable actor-critic architectures: the generated programs specify network structure, while PPO learns the weights through environment interaction. This positions the LLM as a reusable design operator within a meta-evolution loop, rather than as a replacement for RL. Our contributions are as follows:
\begin{itemize}
\item We identify two challenges in automated actor-critic architecture design, namely evaluation cost and the open-ended design space, and formulate a bi-level meta-evolution framework that addresses them jointly.
\item We instantiate architecture search with a low-fidelity PPO evaluator, which provides a practical fitness signal for comparing candidate architectures during evolution.
\item We introduce an LLM design agent as a program-level architecture operator for initialization, mutation, and crossover within an evolutionary loop.
\item Experiments on Ant-v4 and HalfCheetah-v4 show that \evom improves over the manually designed baseline, an LLM-guided random search, and an \mles-style programmatic policy search baseline, with ablations indicating the contributions of evolutionary inheritance and LLM-guided design.
\end{itemize}

\section{Related Work}

\evom is related to three areas: NAS \cite{elsken2019neural} and neuroevolution \cite{miikkulainen2025neuroevolution}, LLM-assisted automated algorithm design \cite{10993463,romera2024mathematical,10.1145/3787585}, and actor-critic reinforcement learning \cite{sutton1998reinforcement,grondman2012survey}. Prior work has studied how to automate architecture design, evolve executable structures, and use LLMs to generate algorithms or programs. In contrast, \evom focuses on LLM-guided evolution of actor-critic architectures, where the generated programs define network structure and PPO remains responsible for policy learning.

\textbf{NAS and neuroevolution.}
NAS automates network design through controller policies, evolutionary search, and differentiable relaxation \cite{zoph2017neural,real2019regularized,elsken2019neural}. Evolutionary computation and genetic programming provide a broader framework for reproduction, evaluation, and selection over executable structures \cite{koza1992genetic,back1996evolutionary}. Neuroevolution applies related principles to policies, weights, and topologies \cite{stanley2002evolving,salimans2017evolution,such2017deep,gaier2019weight}. \evom follows this evolutionary view, but applies LLM-guided program generation to the design of trainable actor-critic architectures.

\textbf{LLM-assisted automated algorithm design.}
LLMs can generate programs, use feedback, and support iterative design loops \cite{brown2020language,ouyang2022training,achiam2023gpt4,chen2021evaluating,yao2023react,shinn2023reflexion,wang2024voyager,wang2025large}. Recent studies use LLMs for automated algorithm design and program search, including mathematical discovery, heuristic design, self-refinement, black-box optimization, and reward design \cite{romera2024mathematical,liu2024eoh,madaan2023self,zhang2023using,ma2024eureka,wang2026taskfree,11493956}. Closely related to our setting, \mles evolves interpretable programmatic controllers with multimodal feedback \cite{hu2025mles}. In contrast, \evom uses the LLM design agent to generate trainable actor-critic architectures, leaving policy and value parameters to be learned by PPO through environment interaction.

\textbf{Actor-critic reinforcement learning.}
Actor-critic methods, including PPO \cite{schulman2017proximal}, learn separate policy and value functions and are widely used for continuous-control tasks. In practice, these functions are instantiated by neural networks whose architecture can influence optimization stability, exploration behavior, and advantage estimation. We use PPO from Stable-Baselines3 \cite{raffin2021stable} on MuJoCo tasks \cite{brockman2016openai,towers2024gymnasium} as the inner learner for evaluating candidate architectures. Because \evom outputs explicit programs, the resulting architectures can be inspected, reproduced, and reused rather than treated as opaque policies.

\section{Problem Formulation}

We consider episodic continuous-control environments with observations $o_t \in \mathbb{R}^d$, actions $a_t \in \mathbb{R}^m$, rewards $r_t$, and horizon $H$. Let $p \in \mathcal{A}$ denote an actor-critic architecture program from an open-ended architecture space $\mathcal{A}$. Rather than assuming a fixed topology or a closed set of hand-specified choices, $\mathcal{A}$ is defined by an executable interface and validity constraints. Given $p$, the actor and critic are instantiated as
\begin{equation}
    \mu_\theta(o) = \pi_\theta(o; p), \qquad v_\phi(o) = V_\phi(o; p),
\end{equation}
where $\mu_\theta(o)$ is the continuous-action mean, $v_\phi(o)$ is a scalar value estimate, and $(\theta,\phi)$ are trainable parameters. Thus, architecture search concerns the outer choice of $p$, while PPO performs the inner optimization of $(\theta,\phi)$ for an architecture.

Given an architecture $p$ and a training seed $s$, let $U_B(p,s)$ denote PPO training under budget $B$. The result of $U_B(p,s)$ is a trained actor-critic policy whose deterministic evaluation return is denoted by
\begin{equation}
    J(U_B(p,s)) =
    \mathbb{E}\left[\sum_{t=0}^{H-1} r_t \mid U_B(p,s)\right].
\end{equation}

Let $\mathcal{A}_{\mathrm{valid}} \subseteq \mathcal{A}$ denote the subset of programs that compile, satisfy the required actor-critic interface, produce tensors with the expected shapes, and return finite outputs. The full-budget architecture search objective is
\begin{equation}
    \max_{p \in \mathcal{A}_{\mathrm{valid}}}
    \; \mathbb{E}_{s}\left[J(U_{B_{\mathrm{full}}}(p,s))\right],
\end{equation}
where $B_{\mathrm{full}}$ denotes the final training budget. Directly optimizing this objective is expensive because each architecture evaluation requires an RL training run. During search, we therefore use a lower training budget $B_{\mathrm{low}} \ll B_{\mathrm{full}}$ to obtain a proxy fitness signal. For an architecture $p$ trained with seed $s$, the low-fidelity estimate is
\begin{equation}
    \hat{J}_{\mathrm{low}}(p,s)
    =
    \frac{1}{K}\sum_{k=1}^{K}
    J_k\left(U_{B_{\mathrm{low}}}(p,s)\right),
\end{equation}
where $J_k$ denotes the return of the $k$-th deterministic evaluation episode. This formulation separates architecture selection from parameter learning: the outer problem searches over architecture programs, while PPO learns the policy and value parameters for each architecture.

\section{EVOM}

\subsection{Overview}

Fig.~\ref{fig:overview} summarizes the \evom workflow. Given the problem setup, the LLM design agent first performs population initialization to produce an initial population $P^{(0)}=\{p_1^{(0)},\ldots,p_N^{(0)}\}$, where $N$ is the population size. The evolution loop then runs for $G$ generations, repeating reproduction, evaluation, ranking, and selection. In reproduction, mutation (edit architectures) and crossover (fuse architectures) generate an offspring population $P'=\{p'_1,\ldots,p'_N\}$. In evaluation, the inner PPO evaluator (low-fidelity) trains each offspring under the low-budget setting and estimates its return using $K$ deterministic evaluation episodes. Ranking then orders the union of the current elite population and the offspring population according to the evaluation signal. Selection keeps the top $N$ candidates as the next elite population $P^{(t+1)}$. After evolution, final evaluation retrains selected elites with full PPO training, and the output stage returns a deployable elite architecture. Algorithm~\ref{alg:evom} describes the workflow. A key design choice is that the LLM design agent does not observe gradients, trained parameter values, or privileged simulator states. It receives the \emph{Environment Description}, \emph{Fixed State-Action} information, \emph{Output Requirements}, and selected architecture designs, and returns executable actor-critic architecture programs. The LLM proposes architecture programs for the outer loop, while PPO training and environment interaction remain in the inner loop.

\begin{algorithm}[t]
\caption{\evom}
\label{alg:evom}
\footnotesize
\begin{algorithmic}[1]
\REQUIRE Environment, observation dimension $d$, action dimension $m$, population size $N$, generations $G$, low-budget $B_{\mathrm{low}}$, full-budget $B_{\mathrm{full}}$.
\ENSURE Best actor-critic architecture and final return $r_{\mathrm{final}}$.
\STATE Use the LLM design agent for population initialization and obtain $P^{(0)}=\{p_1^{(0)},\ldots,p_N^{(0)}\}$;
\STATE Evaluate each architecture $p_i^{(0)}$ with the inner PPO evaluator under budget $B_{\mathrm{low}}$;
\FOR{$t=0,\ldots,G-1$}
    \STATE Generate offspring $P'=\{p'_1,\ldots,p'_N\}$ with the LLM design agent using mutation or crossover over $P^{(t)}$;
    \FOR{each architecture $p'_i \in P'$}
        \STATE Train the policy and value networks defined by $p'_i$ with PPO under budget $B_{\mathrm{low}}$;
        \STATE Estimate the evaluation signal $r_i$ of $p'_i$ over $K$ deterministic evaluation episodes;
    \ENDFOR
    \STATE Rank candidates in $P^{(t)} \cup P'$ by their evaluation signals;
    \STATE Select the top $N$ candidates as the next elite population $P^{(t+1)}$;
\ENDFOR
\STATE Retrain the best architecture in $P^{(G)}$ under budget $B_{\mathrm{full}}$, evaluate it, and return $r_{\mathrm{final}}$;
\end{algorithmic}
\end{algorithm}

\subsection{LLM-Guided Evolution Loop}

The evolution process begins with population initialization. The LLM design agent receives the task description, input and output dimensions, program interface, and design constraints, and returns the initial population $P^{(0)}$. Each architecture $p_i^{(0)}$ is evaluated immediately, so the initial elite population is selected according to empirical feedback rather than model preference alone. For each generation $t$, the outer loop constructs a prompt context containing the \emph{Environment Description}, \emph{Fixed State-Action} information, \emph{Output Requirements}, selected architecture designs, and an evolution directive. The evolution directive asks the agent to perform either mutation on one program or crossover on two programs. The LLM design agent then emits the offspring population $P'$. Each offspring $p'_i$ in $P'$ is compiled, trained with PPO under $B_{\mathrm{low}}$, evaluated, and then ranked together with the current elite population for selection.

The loop is agentic in a restricted sense: it maintains state, takes design actions, and observes empirical feedback. The state is the current elite population together with historical evaluation logs, the actions are program-generation steps for mutation or crossover, and the observation is the evaluation signal returned by the inner PPO evaluator. Because each prompt is conditioned on the current elite population, \evom can reuse design patterns across generations rather than drawing independent LLM samples at each step.

The prompt state is intentionally compact. For mutation, the prompt records \emph{Current Architecture} with its label and network design; for crossover, it records \emph{Parent 1} and \emph{Parent 2} with their labels and network designs. This gives the LLM design agent enough context to compare design patterns without exposing raw trajectories, gradients, or training internals. This compact state representation also supports controlled ablations: removing crossover, removing mutation, or replacing the LLM backbone changes the design operator while keeping the inner PPO evaluator and selection rule fixed.

\subsection{Open-Ended Architecture Representation}
\label{subsec:architecture-interface}

Each $p_i^{(t)}$ is represented as an actor-critic architecture program, which allows the search to evolve architectures beyond a fixed topology or a closed set of layer choices. The program representation remains bounded by a common interface, making open-ended evolution compatible with systematic evaluation. Each program contains a generation label, a structured design description, and an executable Python program. The description records layer types, widths, activations, normalization choices, residual structure, initialization, and actor-critic sharing or asymmetry. The executable program is the authoritative object used for evaluation.

The required interface follows the actor-critic architecture output in Fig.~\ref{fig:overview}. Each $p_i^{(t)}$ must define \texttt{PolicyNet(obs\_length, act\_length)}, which returns an action mean of shape $(n_b,m)$, and \texttt{ValueNet(obs\_length)}, which returns a value tensor of shape $(n_b,1)$, where $n_b$ is the batch size. Depth, width, activation functions, normalization, residual blocks, dropout, initialization, and actor-critic asymmetry remain open to search. Each program is written to an isolated file, imported by a worker, instantiated with task-specific dimensions, and connected to a custom Stable-Baselines3 actor-critic policy wrapper. Programs that fail to compile, produce non-finite outputs, violate tensor-shape constraints, or crash during training are assigned a penalty fitness of $-1000$. This penalty safeguards PPO evaluation and makes invalid architectures unlikely to survive elitist selection or be used as parents in later generations. This validation step is important because executable-program search is less restricted than a hand-built grammar, and the empirical evaluation loop requires well-defined candidates.

\subsection{LLM-Guided Design Operators}

The outer loop uses an LLM as a reusable design agent rather than using it for one-shot architecture generation. The agent receives elite programs, an evolution directive, and an environment description, and returns a structured description and executable program for each offspring architecture $p'_i$. Population initialization produces the initial set of $N$ programs. Mutation edits one parent program through semantic changes to components such as activations, capacity, normalization, and residual blocks. Crossover combines two parent programs into a single offspring program, rather than mechanically splicing program text. The default schedule uses mixed reproduction, with $N/2$ offspring programs generated by mutation and $N/2$ generated by crossover.

\subsection{PPO Evaluation and Final Retraining}

For each candidate architecture, the \texttt{PolicyNet} and \texttt{ValueNet} are trained with PPO under the low-budget setting $B_{\mathrm{low}}$. The evaluation signal is the mean return over $K$ deterministic evaluation episodes. After evolution, the best architecture is retrained from scratch under the full-budget setting $B_{\mathrm{full}}$. Low-fidelity evaluation guides the evolutionary search, while full-budget PPO training determines the reported performance.

\begin{figure}[!t]
\centering
\includegraphics[width=\columnwidth]{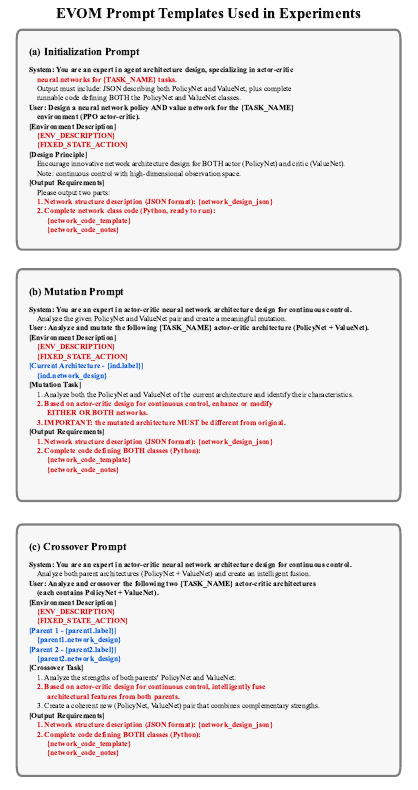}
\caption{Prompt templates for initialization, mutation, and crossover in the \evom design agent. Red placeholders denote fixed task/interface fields, while blue placeholders denote architecture inputs from the current prompt context.}
\label{fig:prompt-templates}
\end{figure}

\section{Experiments}

\subsection{Experimental Setup and Details}

We evaluate \evom on two MuJoCo continuous-control tasks: \textbf{Ant-v4}, with 27-dimensional observations and 8-dimensional continuous torque actions, and \textbf{HalfCheetah-v4}, with 17-dimensional observations and 6-dimensional continuous torque actions. Both tasks use \texttt{VecNormalize}. 

For \evom, the evolutionary search uses population size $N=16$ and $G=20$ generations. For each candidate architecture, the low-budget setting $B_{\mathrm{low}}$ trains PPO for 100k timesteps and estimates the evaluation signal using $K=3$ deterministic evaluation episodes. The 100k-step budget is used as a search-time proxy to indicate whether an actor-critic architecture can be optimized by PPO, while keeping population-level evaluation feasible on consumer GPUs. The full-budget setting $B_{\mathrm{full}}$ trains PPO for 5M timesteps, matching the manually designed baseline budget, and we report final results over three independent runs. The code of \evom can be accessed at \url{https://github.com/xiaofangxd/EVOM}.

As shown in Fig.~\ref{fig:prompt-templates}, \evom uses three prompt templates for initialization, mutation, and crossover. Each prompt provides the environment description, fixed state-action information, and output requirements, including network-design JSON and executable \texttt{PolicyNet}/\texttt{ValueNet} program templates. Initialization creates a new actor-critic architecture from the environment description. Mutation supplies the current architecture label and design, then asks the agent to modify either or both networks while producing a different design. Crossover supplies two parent labels and designs, and asks the agent to fuse architectural features into a coherent new actor-critic program. Each output includes a structured architecture description and a runnable Python program.

\subsection{Baselines and Ablations}

\textbf{Manually designed baseline (Manual PPO)} uses Stable-Baselines3 PPO with a fixed MLP actor-critic architecture \cite{raffin2021stable}. \textbf{Random search} uses the same LLM generation, PPO evaluation, logging, program interface, and final retraining as \evom, but samples new architectures without mutation, crossover, and selection. \textbf{\mles} \cite{hu2025mles} evolves executable controller programs, testing direct programmatic control rather than trainable actor-critic architecture design. For \mles, we use the hyperparameters reported in the original paper. For \evom, we conduct ablations on the reproduction operator, population size, and LLM backbone. These ablations compare mixed reproduction with mutation-only and crossover-only variants, evaluate $N \in \{8,16,32\}$, and test DeepSeek V4, Claude Opus 4.7, and Qwen3.6 Plus.

\subsection{Evaluation Metrics and Visualization}

\begin{figure*}[!t]
\centering
\includegraphics[width=\textwidth]{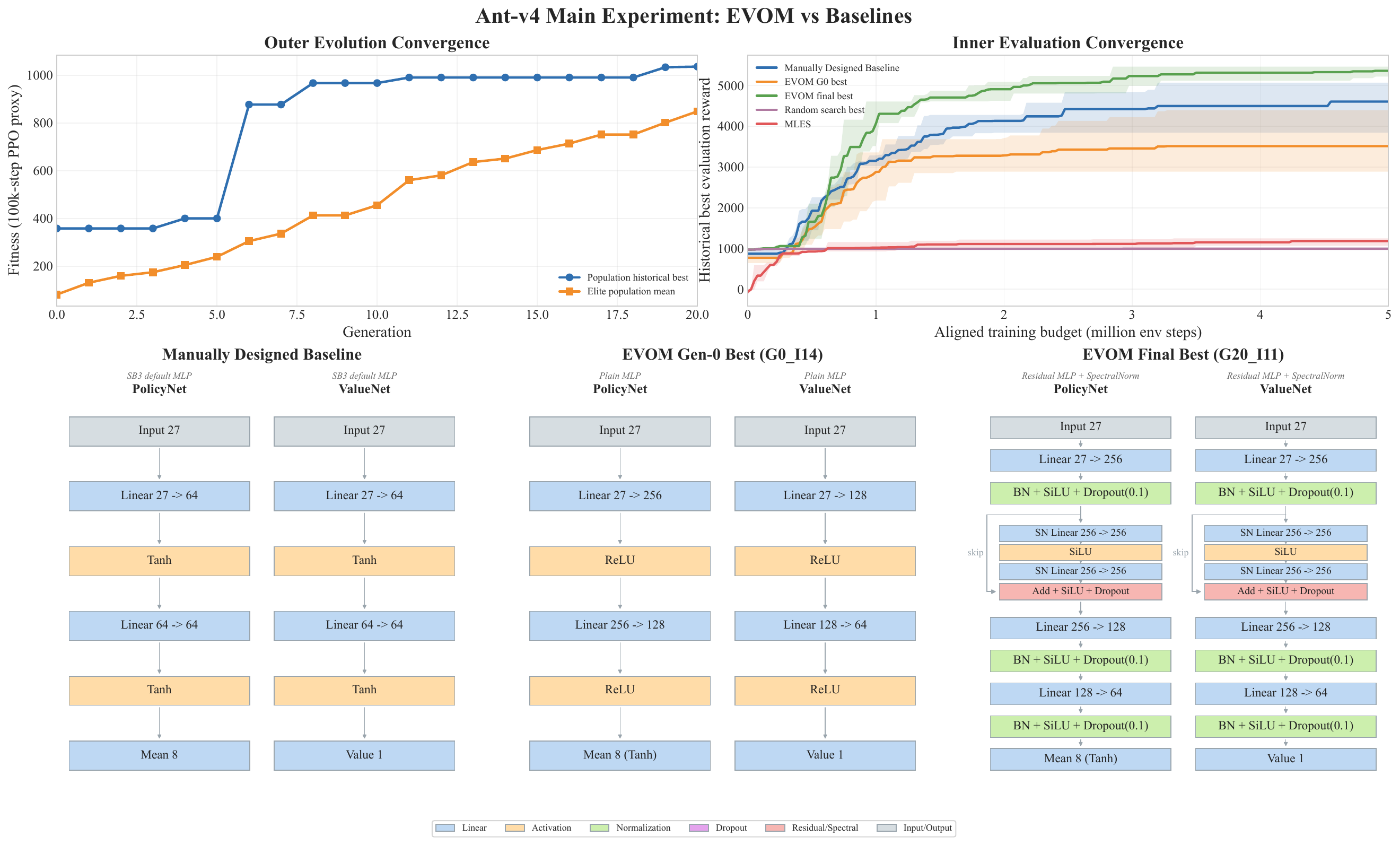}
\caption{Main experimental results on Ant-v4. The evolution loop improves the historical-best reward and the elite-population mean. The final selected actor-critic architecture uses residual and normalization components, differing from the manually designed architecture and the best initial-generation architecture.}
\label{fig:ant-results}
\end{figure*}

\begin{figure*}[!t]
\centering
\includegraphics[width=\textwidth]{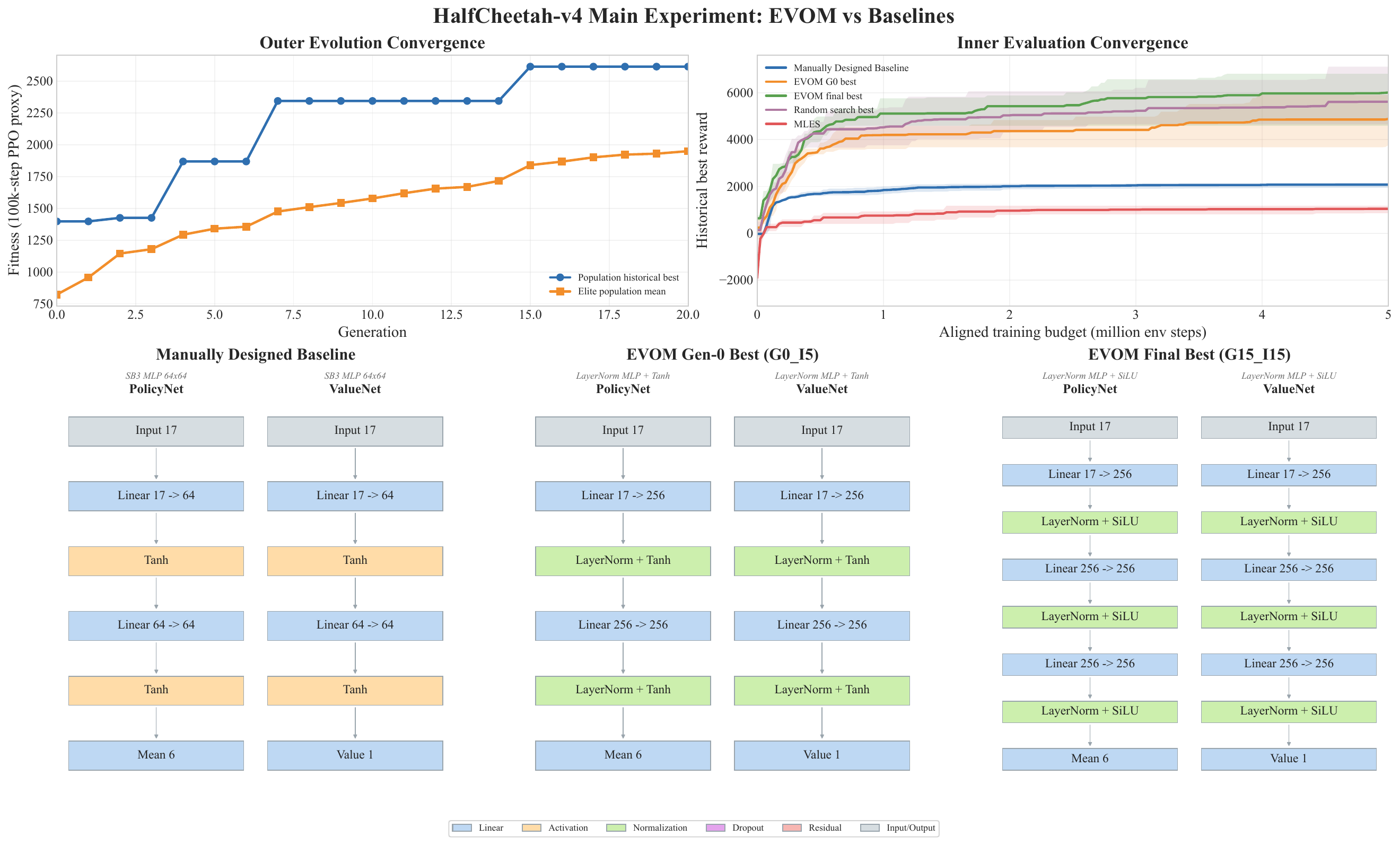}
\caption{Main experimental results on HalfCheetah-v4. The final selected actor-critic architecture uses a deeper LayerNorm-SiLU MLP than the manually designed architecture, while the convergence plots summarize the evolution loop and aligned PPO learning curves under the same full-budget scale.}
\label{fig:half-results}
\end{figure*}

For learning curves, we report the historical-best reward. PPO-based methods use logged evaluation returns, while \mles uses controller-level evaluation logs aligned to the same 5M-step horizontal axis. The architecture diagrams in Figs.~\ref{fig:ant-results} and~\ref{fig:half-results} compare the manually designed architecture, the best initial-generation actor-critic architecture, and the final selected actor-critic architecture, highlighting design patterns such as residual blocks, normalization placement, actor-critic asymmetry, and output-scale handling.

\section{Experimental Results}

\subsection{Full-Budget Performance}

Table~\ref{tab:main} reports full-budget performance after final retraining. On Ant-v4, \evom achieves a higher mean return, suggesting that the low-budget evaluation signal can guide the evolution loop toward effective actor-critic architectures. On HalfCheetah-v4, \evom also obtains the highest mean return. Random search is a stronger baseline on HalfCheetah-v4 than on Ant-v4, but its large standard deviation indicates that independent LLM sampling can occasionally find useful architectures while remaining unstable across full-budget retraining runs.

\subsection{Convergence and Architecture Analysis}

Figs.~\ref{fig:ant-results} and~\ref{fig:half-results} summarize evolution-loop convergence, PPO learning curves, and discovered architectures. Ant-v4 shows a clear improvement pattern: both the historical-best reward and the elite-population mean increase over generations, and the final selected actor-critic architecture includes residual and normalization components that are absent from the manually designed baseline. On HalfCheetah-v4, the random-search learning curve reaches high historical-best rewards but has much larger final-evaluation variance, whereas \evom gives a higher and more reliable full-budget mean. This gap between historical-best curves and final retraining results shows why the tables report final full-budget evaluation rather than search-time or best-checkpoint rewards.

\subsection{Operator and Population Ablations}

Table~\ref{tab:ablations} reports ablations on the reproduction operator and population size. Mixed reproduction gives the best mean return on both tasks. The crossover-only variant remains competitive on Ant-v4 and improves over mutation-only on HalfCheetah-v4, but both single-operator variants underperform mixed \evom. This suggests that local editing and design-pattern recombination are complementary. Increasing the population size does not lead to monotonic gains, likely because larger candidate pools also increase reliance on noisy low-budget evaluation signals.

\begin{table}[t]
\centering
\footnotesize
\begin{tabular}{@{}lcc@{}}
\toprule
Method & Ant-v4 & HalfCheetah-v4 \\
\midrule
Manual PPO & $2517.06 \pm 366.24$ & $1691.42 \pm 495.26$ \\
\mles & $1185.26 \pm 84.15$ & $1045.71 \pm 134.04$ \\
Random Search & $988.15 \pm 2.81$ & $3612.87 \pm 1953.45$ \\
\evom & $4652.10 \pm 746.91$ & $5381.07 \pm 1357.70$ \\
\bottomrule
\end{tabular}
\caption{Main comparison under full-budget evaluation. Values are mean reward $\pm$ standard deviation.}
\label{tab:main}
\end{table}

\begin{table}[t]
\centering
\footnotesize
\setlength{\tabcolsep}{2pt}
\begin{tabular}{@{}lcc@{}}
\toprule
Variant & Ant-v4 & HalfCheetah-v4 \\
\midrule
Mixed EVOM ($N=16$) & $4652.10 \pm 746.91$ & $5381.07 \pm 1357.70$ \\
Mutation only & $2375.43 \pm 1289.72$ & $3455.77 \pm 2030.79$ \\
Crossover only & $4520.79 \pm 486.81$ & $4185.31 \pm 905.67$ \\
Population $N=8$ & $2236.30 \pm 972.23$ & $2340.89 \pm 1379.89$ \\
Population $N=32$ & $2987.88 \pm 1082.51$ & $2797.61 \pm 304.37$ \\
\bottomrule
\end{tabular}
\caption{Ablation results under full-budget evaluation. Values are mean reward $\pm$ standard deviation.}
\label{tab:ablations}
\end{table}

\subsection{LLM Backbone Comparison}

Table~\ref{tab:llm} compares different LLM backbones. Among the tested backbones, Claude Opus 4.7 yields the highest mean return on Ant-v4, whereas DeepSeek V4 yields the highest mean return on HalfCheetah-v4. The strong but task-dependent results of Claude Opus 4.7 and Qwen3.6 Plus indicate that the LLM backbone changes the distribution of generated architectures, while the PPO evaluator and selection loop remain necessary to identify which designs survive full-budget training.

\begin{table}[t]
\centering
\footnotesize
\begin{tabular}{@{}lcc@{}}
\toprule
LLM backbone & Ant-v4 & HalfCheetah-v4 \\
\midrule
DeepSeek V4 & $4652.10 \pm 746.91$ & $5381.07 \pm 1357.70$ \\
Claude Opus 4.7 & $4720.31 \pm 769.87$ & $4445.80 \pm 737.55$ \\
Qwen3.6 Plus & $3964.66 \pm 1127.64$ & $4207.21 \pm 1023.14$ \\
\bottomrule
\end{tabular}
\caption{LLM backbone comparison for the design agent. Values are mean reward $\pm$ standard deviation.}
\label{tab:llm}
\end{table}

\subsection{Architecture-Level Analysis}

The selected elite architectures differ from the manual PPO architecture in several recurring design choices. The Ant-v4 elite uses residual blocks, normalization, SiLU activations, dropout, and output-scale handling, while the HalfCheetah-v4 elite uses a deeper LayerNorm-SiLU MLP with a larger hidden width than the manually designed 64-by-64 MLP. These design patterns are relevant to PPO because the policy network affects action means and exploration behavior, while the value network affects advantage estimation and policy-update quality.

The best initial-generation architectures serve as controls for one-shot LLM design. In both tasks, the final selected architectures differ from these initial designs, especially in residual organization and normalization placement. This suggests that \evom does more than select the best initial sample, as the evolution loop preserves and recombines design patterns across generations. Program-level search also provides a practical benefit: each $p_i^{(t)}$ is an executable actor-critic architecture program, so the final design can be inspected, reproduced, and reused. This contrasts with using the LLM only as a natural-language advisor.

\section{Discussion and Limitations}

\evom uses low-budget PPO evaluation to make population-level search feasible. The resulting proxy can rank many candidate architectures efficiently, but it may not reliably predict full-budget performance. In particular, 100k-step returns may favor architectures that learn quickly but plateau early, or underestimate architectures that require longer training. We therefore treat evolution curves as search diagnostics and report final results after retraining selected elites under the 5M-step full-budget setting.

The benefit of architecture evolution is consistent across the two evaluated tasks. On Ant-v4, \evom substantially improves over both the manually designed baseline and random search, suggesting that evolutionary refinement helps discover architectures that are not easily obtained by independent sampling. On HalfCheetah-v4, \evom also outperforms the compared baselines. These results suggest that meta-evolution can improve architecture discovery beyond one-shot or independent LLM-generated candidates.

The LLM design agent may generate programs that are invalid, unstable under PPO, or unnecessarily expensive to train. We assign penalty fitness to failed programs, but future work should reduce such wasted evaluations through program repair or adaptive evaluation budgets.

The current experiments are limited to two MuJoCo locomotion tasks. A broader evaluation is needed to assess generality, especially on tasks with different observation structures, reward densities, contact dynamics, or action spaces. Future work should also study the transfer of discovered architecture patterns across related tasks and coordination with PPO hyperparameter tuning.

\section{Conclusion}

We presented \evom, an agentic meta-evolution framework for automated actor-critic architecture search. \evom uses an open-ended program representation, LLM-guided design operators, and low-fidelity PPO evaluation to refine candidate architectures across generations. Experiments on Ant-v4 and HalfCheetah-v4 show that \evom discovers actor-critic architectures that outperform the manually designed PPO architecture, \mles-style direct policy search, and LLM-guided random search under full-budget evaluation. These results suggest that LLMs can serve as architectural design operators within evolution loops, while policy and value parameters are learned via RL.

\clearpage
\bibliographystyle{named}
\bibliography{references}

@article{lecun2015deep,
  title={Deep learning},
  author={LeCun, Yann and Bengio, Yoshua and Hinton, Geoffrey},
  journal={nature},
  volume={521},
  number={7553},
  pages={436--444},
  year={2015},
  publisher={Nature Publishing Group UK London}
}

@article{liu2021survey,
  title={A survey on evolutionary neural architecture search},
  author={Liu, Yuqiao and Sun, Yanan and Xue, Bing and Zhang, Mengjie and Yen, Gary G and Tan, Kay Chen},
  journal={IEEE transactions on neural networks and learning systems},
  volume={34},
  number={2},
  pages={550--570},
  year={2021},
  publisher={IEEE}
}

@book{sutton1998reinforcement,
  title={Reinforcement learning: An introduction},
  author={Sutton, Richard S and Barto, Andrew G and others},
  year={1998},
  publisher={MIT press Cambridge}
}

@article{li2024bridging,
  title={Bridging evolutionary algorithms and reinforcement learning: A comprehensive survey on hybrid algorithms},
  author={Li, Pengyi and Hao, Jianye and Tang, Hongyao and Fu, Xian and Zhen, Yan and Tang, Ke},
  journal={IEEE Transactions on evolutionary computation},
  year={2024},
  publisher={IEEE}
}

@article{schulman2017proximal,
  title={Proximal policy optimization algorithms},
  author={Schulman, John and Wolski, Filip and Dhariwal, Prafulla and Radford, Alec and Klimov, Oleg},
  journal={arXiv preprint arXiv:1707.06347},
  year={2017}
}

@article{raffin2021stable,
  author = {Antonin Raffin and Ashley Hill and Adam Gleave and Anssi Kanervisto and Maximilian Ernestus and Noah Dormann},
  title = {Stable-Baselines3: Reliable Reinforcement Learning Implementations},
  journal = {Journal of Machine Learning Research},
  volume = {22},
  number = {268},
  pages = {1--8},
  year = {2021}
}

@ARTICLE{11493956,
  author={Wang, Chao and Li, Lingling and Jiao, Licheng and Zhao, Jiaxuan and Liu, Fang and Yang, Shuyuan},
  journal={IEEE Transactions on Pattern Analysis and Machine Intelligence}, 
  title={Learning Evolution Via Optimization Knowledge Adaptation}, 
  year={2026},
  volume={},
  number={},
  pages={1-18},
  doi={10.1109/TPAMI.2026.3686919}}

@inproceedings{
wang2026taskfree,
title={Task-free Adaptive Meta Black-box Optimization},
author={Chao Wang and Licheng Jiao and Lingling Li and Jiaxuan Zhao and Guanchun Wang and Fang Liu and Shuyuan Yang},
booktitle={The Fourteenth International Conference on Learning Representations},
year={2026},
url={https://openreview.net/forum?id=AufVSUgMUo}
}

@article{wang2025large,
  title={When large language models meet evolutionary algorithms: Potential enhancements and challenges},
  author={Wang, Chao and Zhao, Jiaxuan and Jiao, Licheng and Li, Lingling and Liu, Fang and Yang, Shuyuan},
  journal={Research},
  volume={8},
  pages={0646},
  year={2025},
  publisher={AAAS}
}

@inproceedings{
towers2024gymnasium,
title={Gymnasium: A Standard Interface for Reinforcement Learning Environments},
author={Mark Towers and Ariel Kwiatkowski and John U. Balis and Gianluca De Cola and Tristan Deleu and Manuel Goul{\~a}o and Kallinteris Andreas and Markus Krimmel and Arjun KG and Rodrigo De Lazcano Perez-Vicente and J K Terry and Andrea Pierr{\'e} and Sander V Schulhoff and Jun Jet Tai and Hannah Tan and Omar G. Younis},
booktitle={The Thirty-ninth Annual Conference on Neural Information Processing Systems Datasets and Benchmarks Track},
year={2025},
url={https://openreview.net/forum?id=qPMLvJxtPK}
}

@inproceedings{zoph2017neural,
  author = {Barret Zoph and Quoc V. Le},
  title = {Neural Architecture Search with Reinforcement Learning},
  booktitle = {International Conference on Learning Representations},
  year = {2017}
}

@inproceedings{real2019regularized,
  author = {Esteban Real and Alok Aggarwal and Yanping Huang and Quoc V. Le},
  title = {Regularized Evolution for Image Classifier Architecture Search},
  booktitle = {Proceedings of the AAAI Conference on Artificial Intelligence},
  volume = {33},
  pages = {4780--4789},
  year = {2019}
}

@article{elsken2019neural,
  author = {Thomas Elsken and Jan Hendrik Metzen and Frank Hutter},
  title = {Neural Architecture Search: A Survey},
  journal = {Journal of Machine Learning Research},
  volume = {20},
  number = {55},
  pages = {1--21},
  year = {2019}
}

@article{stanley2002evolving,
  author = {Kenneth O. Stanley and Risto Miikkulainen},
  title = {Evolving Neural Networks through Augmenting Topologies},
  journal = {Evolutionary Computation},
  volume = {10},
  number = {2},
  pages = {99--127},
  year = {2002}
}

@article{miikkulainen2025neuroevolution,
  title={Neuroevolution insights into biological neural computation},
  author={Miikkulainen, Risto},
  journal={Science},
  volume={387},
  number={6735},
  pages={eadp7478},
  year={2025},
  publisher={American Association for the Advancement of Science}
}

@ARTICLE{10993463,
  author={Ma, Zeyuan and Guo, Hongshu and Gong, Yue-Jiao and Zhang, Jun and Tan, Kay Chen},
  journal={IEEE Transactions on Evolutionary Computation}, 
  title={Toward Automated Algorithm Design: A Survey and Practical Guide to Meta-Black-Box-Optimization}, 
  year={2026},
  volume={30},
  number={2},
  pages={667-687},
  doi={10.1109/TEVC.2025.3568053}}

@article{romera2024mathematical,
  title={Mathematical discoveries from program search with large language models},
  author={Romera-Paredes, Bernardino and Barekatain, Mohammadamin and Novikov, Alexander and Balog, Matej and Kumar, M Pawan and Dupont, Emilien and Ruiz, Francisco JR and Ellenberg, Jordan S and Wang, Pengming and Fawzi, Omar and others},
  journal={Nature},
  volume={625},
  number={7995},
  pages={468--475},
  year={2024},
  publisher={Nature Publishing Group UK London}
}

@article{grondman2012survey,
  title={A survey of actor-critic reinforcement learning: Standard and natural policy gradients},
  author={Grondman, Ivo and Busoniu, Lucian and Lopes, Gabriel AD and Babuska, Robert},
  journal={IEEE Transactions on Systems, Man, and Cybernetics, part C (applications and reviews)},
  volume={42},
  number={6},
  pages={1291--1307},
  year={2012},
  publisher={IEEE}
}

@article{10.1145/3787585,
author = {Liu, Fei and Yao, Yiming and Guo, Ping and Yang, Zhiyuan and Lin, Xi and Zhao, Zhe and Tong, Xialiang and Mao, Kun and Lu, Zhichao and Wang, Zhenkun and Yuan, Mingxuan and Zhang, Qingfu},
title = {A Systematic Survey on Large Language Models for Algorithm Design},
year = {2026},
issue_date = {June 2026},
publisher = {Association for Computing Machinery},
address = {New York, NY, USA},
volume = {58},
number = {8},
issn = {0360-0300},
url = {https://doi.org/10.1145/3787585},
doi = {10.1145/3787585},
journal = {ACM Comput. Surv.},
month = feb,
articleno = {218},
numpages = {32}
}

@inproceedings{
hu2025mles,
title={Multimodal {LLM}-assisted Evolutionary Search for Programmatic Control Policies},
author={Qinglong Hu and Tong Xialiang and Mingxuan Yuan and Fei Liu and Zhichao Lu and Qingfu Zhang},
booktitle={The Fourteenth International Conference on Learning Representations},
year={2026},
url={https://openreview.net/forum?id=OHFNJoNtjW}
}

@book{koza1992genetic,
  author = {John R. Koza},
  title = {Genetic Programming: On the Programming of Computers by Means of Natural Selection},
  publisher = {MIT Press},
  year = {1992}
}

@book{back1996evolutionary,
  author = {Thomas B{\"a}ck},
  title = {Evolutionary Algorithms in Theory and Practice},
  publisher = {Oxford University Press},
  year = {1996}
}

@techreport{achiam2023gpt4,
  author = {{OpenAI}},
  title = {GPT-4 Technical Report},
  institution = {OpenAI},
  year = {2023},
  eprint = {2303.08774},
  archivePrefix = {arXiv},
  primaryClass = {cs.CL}
}

@article{brockman2016openai,
  title={Openai gym},
  author={Brockman, Greg and Cheung, Vicki and Pettersson, Ludwig and Schneider, Jonas and Schulman, John and Tang, Jie and Zaremba, Wojciech},
  journal={arXiv preprint arXiv:1606.01540},
  year={2016}
}

@inproceedings{yao2023react,
  author = {Shunyu Yao and Jeffrey Zhao and Dian Yu and Nan Du and Izhak Shafran and Karthik Narasimhan and Yuan Cao},
  title = {ReAct: Synergizing Reasoning and Acting in Language Models},
  booktitle = {International Conference on Learning Representations},
  year = {2023}
}

@inproceedings{shinn2023reflexion,
  author = {Noah Shinn and Federico Cassano and Ashwin Gopinath and Karthik Narasimhan and Shunyu Yao},
  title = {Reflexion: Language Agents with Verbal Reinforcement Learning},
  booktitle = {Advances in Neural Information Processing Systems},
  year = {2023}
}

@article{
wang2024voyager,
title={Voyager: An Open-Ended Embodied Agent with Large Language Models},
author={Guanzhi Wang and Yuqi Xie and Yunfan Jiang and Ajay Mandlekar and Chaowei Xiao and Yuke Zhu and Linxi Fan and Anima Anandkumar},
journal={Transactions on Machine Learning Research},
issn={2835-8856},
year={2024},
url={https://openreview.net/forum?id=ehfRiF0R3a},
note={}
}

@inproceedings{
liu2024eoh,
title={Evolution of Heuristics: Towards Efficient Automatic Algorithm Design Using Large Language Model},
author={Fei Liu and Tong Xialiang and Mingxuan Yuan and Xi Lin and Fu Luo and Zhenkun Wang and Zhichao Lu and Qingfu Zhang},
booktitle={Forty-first International Conference on Machine Learning},
year={2024},
url={https://openreview.net/forum?id=BwAkaxqiLB}
}

@inproceedings{ma2024eureka,
  author = {Yecheng Jason Ma and William Liang and Guanzhi Wang and De-An Huang and Osbert Bastani and Dinesh Jayaraman and Yuke Zhu and Linxi Fan and Anima Anandkumar},
  title = {Eureka: Human-Level Reward Design via Coding Large Language Models},
  booktitle = {International Conference on Learning Representations},
  year = {2024}
}

@inproceedings{brown2020language,
  author = {Tom B. Brown and Benjamin Mann and Nick Ryder and Melanie Subbiah and Jared Kaplan and Prafulla Dhariwal and Arvind Neelakantan and Pranav Shyam and Girish Sastry and Amanda Askell and Sandhini Agarwal and Ariel Herbert-Voss and Gretchen Krueger and Tom Henighan and Rewon Child and Aditya Ramesh and Daniel M. Ziegler and Jeffrey Wu and Clemens Winter and Christopher Hesse and Mark Chen and Eric Sigler and Mateusz Litwin and Scott Gray and Benjamin Chess and Jack Clark and Christopher Berner and Sam McCandlish and Alec Radford and Ilya Sutskever and Dario Amodei},
  title = {Language Models are Few-Shot Learners},
  booktitle = {Advances in Neural Information Processing Systems},
  year = {2020}
}

@inproceedings{ouyang2022training,
  author = {Long Ouyang and Jeffrey Wu and Xu Jiang and Diogo Almeida and Carroll Wainwright and Pamela Mishkin and Chong Zhang and Sandhini Agarwal and Katarina Slama and Alex Ray and John Schulman and Jacob Hilton and Fraser Kelton and Luke Miller and Maddie Simens and Amanda Askell and Peter Welinder and Paul Christiano and Jan Leike and Ryan Lowe},
  title = {Training Language Models to Follow Instructions with Human Feedback},
  booktitle = {Advances in Neural Information Processing Systems},
  year = {2022}
}

@article{chen2021evaluating,
  title={Evaluating large language models trained on code},
  author={Chen, Mark and Tworek, Jerry and Jun, Heewoo and Yuan, Qiming and Pinto, Henrique Ponde De Oliveira and Kaplan, Jared and Edwards, Harri and Burda, Yuri and Joseph, Nicholas and Brockman, Greg and others},
  journal={arXiv preprint arXiv:2107.03374},
  year={2021}
}

@article{madaan2023self,
  title={Self-refine: Iterative refinement with self-feedback},
  author={Madaan, Aman and Tandon, Niket and Gupta, Prakhar and Hallinan, Skyler and Gao, Luyu and Wiegreffe, Sarah and Alon, Uri and Dziri, Nouha and Prabhumoye, Shrimai and Yang, Yiming and others},
  journal={Advances in neural information processing systems},
  volume={36},
  pages={46534--46594},
  year={2023}
}

@inproceedings{
zhang2023using,
title={Using Large Language Models for Hyperparameter Optimization},
author={Michael Zhang and Nishkrit Desai and Juhan Bae and Jonathan Lorraine and Jimmy Ba},
booktitle={NeurIPS 2023 Foundation Models for Decision Making Workshop},
year={2023},
url={https://openreview.net/forum?id=FUdZ6HEOre}
}

@article{salimans2017evolution,
  title={Evolution strategies as a scalable alternative to reinforcement learning},
  author={Salimans, Tim and Ho, Jonathan and Chen, Xi and Sidor, Szymon and Sutskever, Ilya},
  journal={arXiv preprint arXiv:1703.03864},
  year={2017}
}

@article{such2017deep,
  title={Deep neuroevolution: Genetic algorithms are a competitive alternative for training deep neural networks for reinforcement learning},
  author={Such, Felipe Petroski and Madhavan, Vashisht and Conti, Edoardo and Lehman, Joel and Stanley, Kenneth O and Clune, Jeff},
  journal={arXiv preprint arXiv:1712.06567},
  year={2017}
}

@inproceedings{gaier2019weight,
  author = {Adam Gaier and David Ha},
  title = {Weight Agnostic Neural Networks},
  booktitle = {Advances in Neural Information Processing Systems},
  year = {2019}
}

\end{document}